# Photorealistic Facial Expression Synthesis by the Conditional Difference Adversarial Autoencoder

Yuqian ZHOU, Bertram Emil SHI
*Department of Electronic and Computer Engineering*
*The Hong Kong University of Science and Technology*
*HKSAR, China*
yzhouas@ust.hk, eebert@ust.hk

*Abstract*—Photorealistic facial expression synthesis from single face image can be widely applied to face recognition, data augmentation for emotion recognition or entertainment. This problem is challenging, in part due to a paucity of labeled facial expression data, making it difficult for algorithms to disambiguate changes due to identity and changes due to expression. In this paper, we propose the conditional difference adversarial autoencoder (CDAAE) for facial expression synthesis. The CDAAE takes a facial image of a previously unseen person and generates an image of that person's face with a target emotion or facial action unit (AU) label. The CDAAE adds a feedforward path to an autoencoder structure connecting low level features at the encoder to features at the corresponding level at the decoder. It handles the problem of disambiguating changes due to identity and changes due to facial expression by learning to generate the difference between low-level features of images of the same person but with different facial expressions. The CDAAE structure can be used to generate novel expressions by combining and interpolating between facial expressions/action units within the training set. Our experimental results demonstrate that the CDAAE can preserve identity information when generating facial expression for unseen subjects more faithfully than previous approaches. This is especially advantageous when training with small databases.

## 1. Introduction

Rendering photorealistic facial expression from a single static face while preserving the identity information will have significant impact in the area of affective computing. Generated faces of a specific person with different facial expressions can be applied to emotion prediction, face recognition, expression database augmentation, entertainment, etc. Although prior works have shown how to transfer facial expressions between subjects, *i.e.* facial reenactment [1], or to synthesize facial expressions on a virtual agent [2], the problem of synthesizing a wide range of facial expressions accurately on arbitrary real faces is still an open problem.

This paper describes a system that takes an arbitrary face image with a random (i.e., not necessarily neutral) facial expression and synthesizes a new face image of the same person, but with a different expression, as defined by an emotion (e.g. happiness, sadness, etc.), or by varying levels of facial action unit (AU) intensity, as defined by the Facial Action Coding System (FACS) [3] (e.g., lip corner up, inner brow up etc.). This work is challenging because databases with labeled facial expressions, like CK+ [4] and DISFA [5], are usually small, containing only about 100 subjects or less. Although the databases contain images with a large variety of facial expressions, because they have so few subjects, it is hard to disentangle facial expression and identity information. Due to this difficulty, prior work has considered the problem of generating expressions only for subjects in the training set. These approaches, based on deep belief nets (DBNs) [6] or deconvolutional neural networks (DeCNNs) [7], essentially generate faces by interpolation among images in the training set, making them inherently unsuited for facial expression generation for unseen subjects.

With the recent development of generative adversarial networks (GANs) [8], image editing has migrated from pixel-level manipulations to semantic-level manipulations. GANs have been successfully applied to face image editing, e.g., age modeling [9], pose adjustment [10] and the modification of facial attributes [11], [12]. These works generally use the encoder of the GAN to find a low-dimensional representation of the face image in a latent space, manipulate the latent vector, and then decode it to generate the new image. Popular manipulations of the latent vector include shifting the latent vector along the specific direction corresponding to semantic attributes by using vector arithmetic [13], [14], or directly concatenating attribute labels with the latent vector [9], [11]. Adversarial discriminator networks are used either at the encoder to regularize the latent space [15], at the decoder to generate blur-free and realistic images [13] or at both the encoder and decoder, i.e., the Conditional Adversarial Autoencoder (CAAE) [9]. All of these approaches require large training databases so that identity information can be properly disambiguated. Otherwise, when presented with an unseen face, the network tends to generate faces which look like the "closest" subject in the training dataset.

Yeh et al. proposed to handle this problem by warping images, rather than generating them from the latent vector directly [16]. This approach captures the idea that facial expressions generally affect small face regions, rather than whole images. A mapping is learned from the latent space to a flow space, which is used to warp the original images. To generate a new facial expression, the latent vector is modified by vector arithmetic: adding the averaged difference between the latent vectors of images with target and source labels. The approach achieves a high quality for interpolation, but

This work was supported in part by the General Research Fund of the Hong Kong Research Grants Council, under grant 618713.

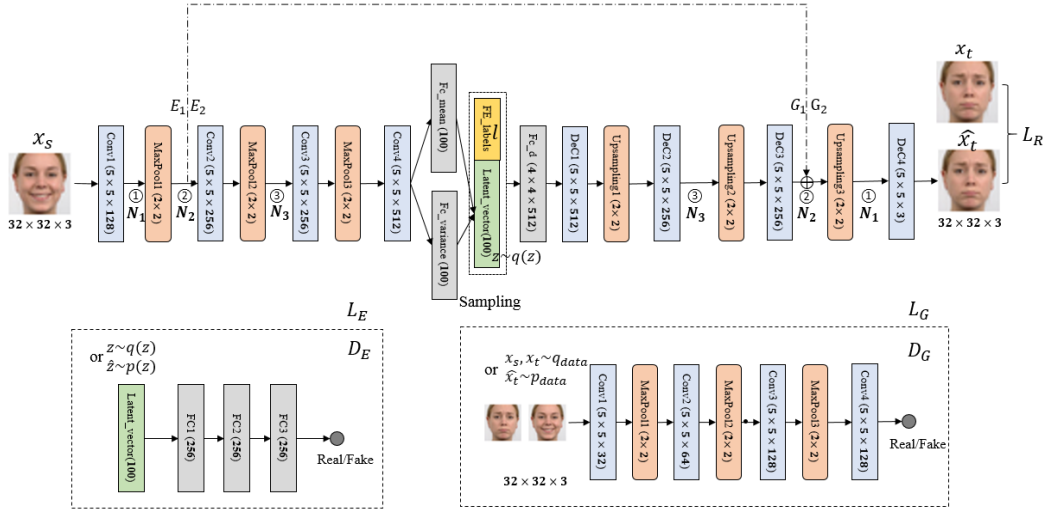

Fig.1. The system structure of the proposed CDAAE. The network takes faces of the source expression and an expression label to generate images of the target expression. The input goes through an encoder $E$ to map the raw pixels to a latent space distribution $q(z)$. The latent vector $z$ and the target label vector $l$ are then concatenated to generate the target faces through a decoder $G$. An additional feedforward connection, shown here as the dashed line connecting position ② in both the encoder and decoder, is added. The connection splits the encoder and decoder into two stages, $E_1, E_2, G_1$ and $G_2$. It forces $E_2$ and $G_1$ to learn and generate the general difference between the source and target expression, and reuse the low-level features computed by $E_1$ for further integrated decoding. Two discriminators, $D_E$ and $D_G$, are imposed on $E_2$ and $G_2$ respectively. $D_E$ is utilized to regularize the latent space distribution $q(z)$ to a Gaussian distribution $p(z)$, and $D_G$ is applied to improve the quality of generated images.

requires that the input expression be known, and fails when mapping between facial expressions that are "far apart," e.g. generating angry faces from smiling faces.

In this paper, we propose an alternative approach to decouple identity and expression information, which works for training databases with limited subjects. We propose the Conditional Difference Adversarial Autoencoder (CDAAE), which augments the adversarial autoencoder with a long-range feedforward connection from the encoder to the decoder. The network models only the changes of low-level facial features conditioned on the desired expression label, rather than the entire face generation process. Specifically, we make two changes to the CAAE [9]. First, instead of utilizing the same images as the input and output during training, we train the network on pairs of images of the same subject, but with different facial expressions. One image is presented at the input, the other at the output. Second, we add a feedforward connection to from a middle layer in the convolutional network to the corresponding layer in the decoder network. We train the model to learn the difference between the feature representations of the source and target images at this level. This enables us to reuse parts of the low-level facial attributes, and to learn the high-level changes of expression. Intuitively, the long range feedforward connection preserves identity information, enabling the rest of the network to focus on modelling changes due to the facial expression.

The proposed CDAAE has two primary contributions. First, it generates accurate expressions for faces unseen in the training set. Because the additional feedforward connection preserves identity information, the network does this even when trained on a database containing only a small number of subjects. Our experimental results demonstrate that the faces generated through the CDAAE are perceptually more similar to the input subject's than faces generated by the CAAE. Second, compared to prior methods, with the CDAAE it is easier to manipulate the generated facial expressions, and even to generate images corresponding to new combinations of facial muscle movements on new subjects.

## 1. Related Work

### 1.1. Facial Expression Synthesis

Facial expression synthesis can be categorized into virtual avatar animation and photorealistic face rendering. The realization of facial expression generation on new subjects can be achieved by expression transfer, which extracts geometric features (facial landmark points) [17], appearance features (wrinkles) [18] or 3D meshes adopted from the RGB-D space [1][19] from images of existing subjects, and maps them to avatars or new faces. Another possibility is cross fading or warping/morphing existing faces based on an existing source. Mohammed et al. [20] composited local patches in the training set to create novel faces with modified facial expressions. Yang et al. [21] proposed an interactive platform that used user-marked facial feature points and default mesh parameters to generate facial expressions on new faces. However, these methods cannot generate expressions based on high-level descriptions like facial expression or AU labels. In addition, the full complexity of natural human expressions and identity-specific characteristics are hard to model only by morphing or other pixel-wise operation.

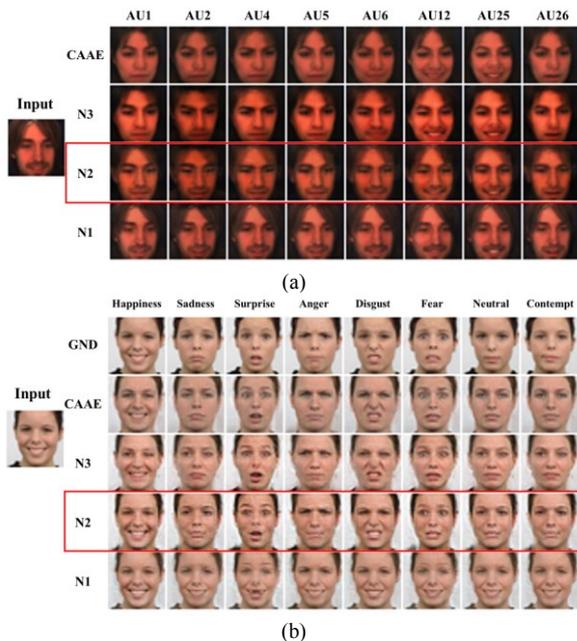

Fig. 2. Comparison of expression generation ability and the identity preserving quality of the four network structures: (a) spontaneous facial expression synthesis on DISFA database and (b) posed expression synthesis on RaFD database. For both types of expression synthesis, N2 not only generates plausible facial expressions, but also simultaneously preserves identity information.

Neural networks now provide better flexibility in image generation. The development of deep networks enables the generation to be guided by semantic labels. Nejadgholi et al. [22] proposed a brain-inspired model to generate prototypes and ambitious exemplars from trained or new neutral faces. However, it cannot model the transformation of expression. Yeh et al. [16] used a variational autoencoder to learn a flow map that could be used to warp source images with one facial expression to target images with a different label, and then applied the flow map to warp images of new faces. These methods attempt to manipulate facial expression synthesis with labels, but do not systematically model the intricate correlations between different expressions and facial movements.

### 1.2. Generative Adversarial Network

The Generative Adversarial Network (GAN) [8] and Deep Convoluntional GAN (DCGAN) [23], establish a min-max two-player adversarial game between a generative and a discriminative model. However, during generation, the latent variable is randomly sampled from a prior distribution, so there is not much control over the output. There are normally two ways to resolve the problem. First, the GAN can be extended to a conditional model called the CGAN [24] by adding auxiliary information like labels. Second, an autoencoder-like structure, such as the variational autoencoder (VAE) [25] or the adversarial autoencoder (AAE) [15], can be used to impose a prior distribution on the GAN. These approaches encode the original data at the pixel level using a low dimensional representation, and to generate novel images from points in this low dimensional subspace. The advantages of the AE in forming a controllable latent space with input, and the benefits of the GAN in estimating the pixel space directly can be combined to manipulate photorealistic images. Zhang et al. [9] integrated both the AAE and the CGAN for age modeling. They proposed a Conditional AAE (CAAE) to learn the face manifold conditioned on age. The success of this model relies upon the availability of a large database with thousands of subjects at different ages, so that identity manifolds can be modeled properly. Unfortunately, it is hard to collect and label facial expression databases with comparable size. Deton et al. [26] proposed a pyramid GAN to generate samples following a coarse-to-fine strategy. This work inspired us to modify the CAAE structure to use the upper (coarse) layers of the AE to learn differences in facial expression, and to use these differences to modify information at a lower (finer) layer.

## 2. Methodology

### 2.1. Datasets

Facial expression database can be categorized as containing either posed or spontaneous expressions. Posed facial expression databases often have extreme facial changes, and are mostly labeled with emotions, e.g., happiness, sadness. Spontaneous facial expressions exhibit more subtle facial changes, which are difficult to classify using only a small number of emotion classes. Instead, these are usually labeled by estimating the intensities of the active facial action units (AUs). In this paper, we consider facial expression synthesis for both posed and spontaneous facial expressions.

The Denver Intensity of Spontaneous Facial Action (DISFA) dataset [5] contains stimulated spontaneous facial expression videos from 27 adult subjects (12 female and 15 male). Each subject appears in only one video, which is recorded as they watch a four-minute emotive video stimulus. Each video contains 4845 frames, and each frame is labeled with the intensity over 12 facial AUs, which are AU1, AU2, AU4, AU5, AU6, AU9, AU12, AU15, AU17, AU20, AU25, and AU26, ranging from zero to five. The range of the labels was rescaled to 0-1 for training.

The Radboud Faces database (RaFD) [27] is a multi-view posed facial expression collection consisting of images of 67 subjects labeled with eight expressions and three gaze directions taken from five viewing angles. In this paper, we used only the frontal-view images with three different gaze directions. We obtained in total 1608 samples ($67 \times 3$ for each class), and conducted a four-fold subject-independent cross validation, by randomly splitting the folds of 16 or 17 subjects.

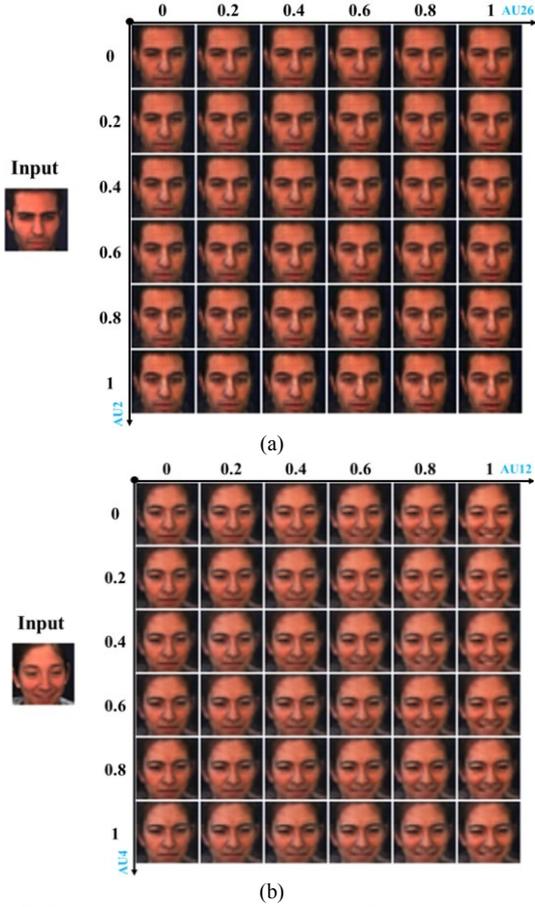

Fig. 3. Example subspaces of the face manifold learned by the N2 structure on the DISFA database. Given an input face with an arbitrary expression, the figure shows (a) generated faces with increasing intensity of AU2 and AU26 and (b) generated faces with increasing intensities of AU4 and AU12.

For each dataset, we aligned the faces by Procrustes analysis using the 68 facial landmarks detected by the Dlib Regression Trees algorithm [28], [29], and cropped them. Each image was further resized to $32 \times 32$.

### 2.2. System Architecture

Fig. 1. demonstrates the detailed structure of the CDAAE. The network takes $32 \times 32$ RGB faces $x \in \mathbf{R}^{32 \times 32 \times 3}$ of the source expression and outputs RGB images $\hat{x} \in \mathbf{R}^{32 \times 32 \times 3}$ of the target expression. The input images first go through a four-layer convolutional encoder $E$, which maps the raw face images to a regularized latent space $q(z)$. The latent vector $z = E(x) \in \mathbf{R}^{100}$ and the target label vector $l$ are then concatenated to generate the target faces through a four-layer deconvolutional decoder $G$. The activation function between each layer is Leaky ReLU with gradient 0.2.

Unlike the traditional autoencoder structure or CAAE, the CDAAE has an additional feedforward connection, which is shown as the dashed line connecting position ② in both the encoder and decoder of Fig. 1. In our experiments, we considered adding this single feedforward connection between points at positions ①, ② and ③, and denoted the networks by $N_1$, $N_2$, and $N_3$. The feedforward connection splits both the encoder and decoder into two stages: $E_1$ and $E_2$ for the encoder, and $G_1$ and $G_2$ for the decoder. This forces the high-level parts of the network, (stages $E_2$ and $G_1$) to learn the difference between the source and target expression: $d = G_1(z, l)$, where $z = E_2(E_1(x_s))$. It also enables the low-level features computed during the encoding process $E_1(x_s)$ to be reused during the decoding process. At the layer learning the difference, we use the tanh activation function instead of Leaky ReLU. Intuitively, most expression-unrelated facial attributes are represented by low-level features and can be reused to maintain the identity information. Finally, the output faces conditioned on specific target expression labels can be expressed as

$$\hat{x}_t = G_2(E_1(x_s) + d)$$
$$= G_2(E_1(x_s) + G_1(E_2(E_1(x_s)), l)) \quad (1)$$

In addition, two discriminators, $D_E$ and $D_G$, are applied to $E_2$ and $G_2$ respectively. $D_E$ is utilized to regularize the latent space $q(z)$ to a Gaussian distribution $p(z)$. $D_G$ is applied to improve the quality of the generated images. The detailed structures of the discriminators are illustrated in Fig. 1.

The training process can be modeled by a min-max objective function

$$\min_{E_1, E_2, G_1, G_2} \max_{D_E, D_G} \alpha L_R + \beta_1 L_E + \beta_2 L_G, \quad (2)$$

where $L_R$ indicates the mean square reconstruction error with the target images, and $L_E$ and $L_G$ are the objective loss for the adversarial process of the encoder and decoder respectively. Specifically,

$$L_R = L_2(x_t, \hat{x}_t) \quad (3)$$

$$L_E = \mathbf{E}_{z^* \sim p(z)}[\log D_E(z^*)]$$
$$+ \mathbf{E}_{x_s \sim p_{data}}[\log(1 - D_E(z))] \quad (4)$$

$$L_G = \mathbf{E}_{x_s \sim p_{data}}[\log D_G(x_s)]$$
$$+ \mathbf{E}_{x_s \sim p_{data}, l \sim p_l}[\log(1 - D_G(\hat{x}_t))] \quad (5)$$

## 3. Experimental Results

### 3.1. Spontaneous Expression Synthesis

We conducted spontaneous expression synthesis experiment on the DISFA database. We used four-fold subject-independent cross validation, by randomly splitting the data into folds containing 6 or 7 subjects. For each training fold and for each AU, we chose as target faces up to 2000 non-zero frames (depending upon the amount of data) following the original intensity label distribution and an additional 1000 zero frames. In total, each fold contained about 30000 target face frames. We paired each target frame with a source frame

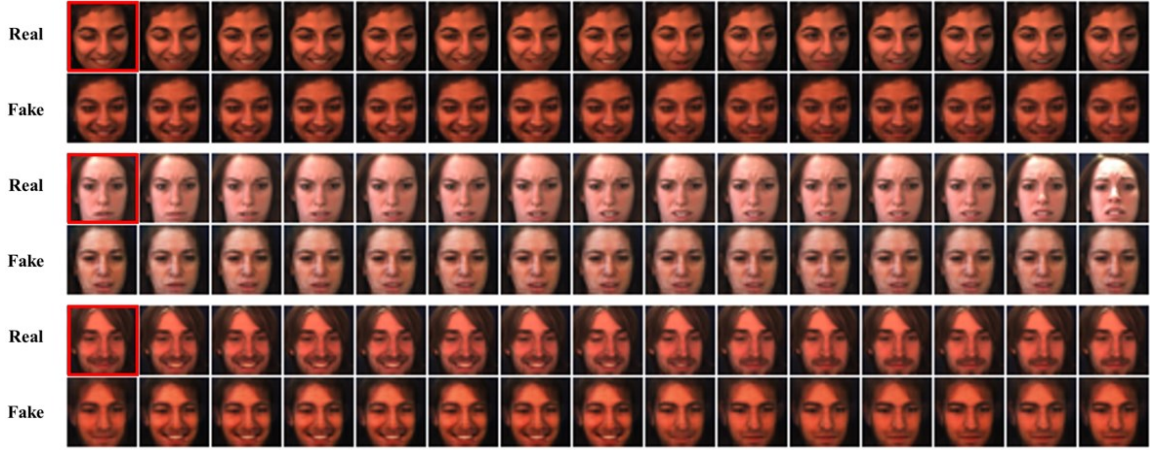

Fig.4. A comparison of real facial images and generated images by the network N2 with the same expression label. We show results from three different subjects. For each subject, we show two rows of images. The top row shows actual images taken from the DISFA database. The bottom row shows images generated by the network whose input is the image framed in red and with the labels set to the corresponding labels in the top row image.

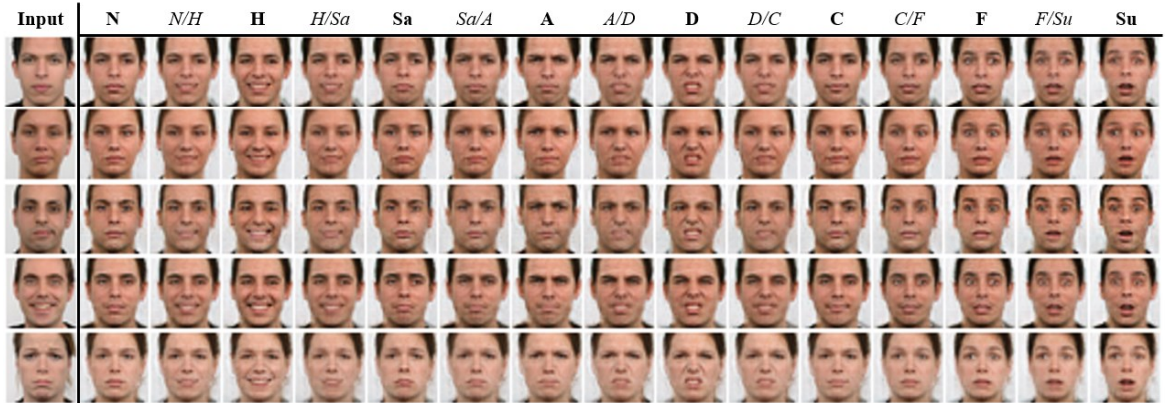

Fig.5. Synthesis results of all the emotion classes and their interpolation using the CDAAE (N2) network. The interpolation is obtained by setting the label values of the two emotions to 0.5. (**N**:neutral, **H**:happiness, **Sa**:sadness, **A**:anger, **D**:disgust, **C**:contempt, **F**:fear, **Su**:surprise)

chosen randomly from the set of facial images of the same person in the target frame.

Fig. 1 presents the network structure. The facial expression labels $l \in \mathbf{R}^{12}$ are represented by the intensities (from zero to one) of the 12 facial AUs. We implemented four network structures: $N_1$, $N_2$, $N_3$ and a network without the long-range feedforward connection, which is similar to the CAAE and is used as a control. We trained the four network settings using the Adam optimizer with learning rate $10^{-3}$ and $10^{-4}$ for the autoencoder and discriminators respectively. The batch size is 32. The discriminators and autoencoder are updated alternately, but the two discriminators ($D_E$ and $D_G$) are trained simultaneously. During training, $\alpha$ is set to 1, and $\beta_1 = 10^{-2}$, and $\beta_2 = 10^{-3}$ empirically. Training is stopped after about 40 epochs until the network generates plausible images. The system is implemented by Keras [30] with a Theano backend [31].

We compare the expression generation ability and the identity preserving quality of the four networks in Fig. 2 (a). Each column shows the generated faces for eight different AUs set to the highest intensity. Each row shows the results generated by a different network structure. The results demonstrate that it is hard to generate facial images of arbitrary subjects using the CAAE structure when it is trained on small databases. With the additional feedforward connection, important facial attributes encoding identity are preserved. Identity preservation improves for networks where the feedforward path connects lower-level features, i.e. the network $N_1$ better preserves identity than the network $N_3$. On the other hand, the expression generation ability is limited with the lower-level connections. The network $N_2$ achieves the best tradeoff between better expression generation and better identity preservation.

Fig. 3 shows the manifold learned by $N_2$ for different combinations of AUs. The input faces were unseen during training. The targeted AU intensity values ranged from zero to one. Fig. 3(a) shows combined expressions generated by simultaneously varying the intensities of AU2 (Outer Brow Raiser) and AU26 (Jaw Drop). The other AU intensity values were set to zero. Similarly, in Fig. 3(b), we only set the intensity values of AU4 (Brow Lower) and AU12 (Lip Corner Puller). The generated faces in these figures show a gradual and smooth transition along the indicated direction of the AU intensity changes. Identity information is greatly preserved, despite dramatic changes of facial expression.

Fig. 4 compares real expressions with corresponding expressions generated by network N2. Our model works well with multi-label inputs, with the generated images being very similar to the actual images, both in terms of expression and identity.

### 3.2. Posed Expression Synthesis

A posed expression synthesis experiment was conducted on the RaFD database. The dataset contained 50 subjects in the training set and 17 subjects in the testing set. We split the training set into folds of 16 to 17 subjects, and conducted four-fold cross validation. The training set contained 1200 images (50 subjects, 3 gaze directions and 8 emotions). We created 9600 source-target pairs by pairing each images in the training set with the 8 images of the same subject and with the same gaze angle by different emotions. This forced the network to model all possible transformations between emotion classes.

The network construction is shown in Fig.1. For the facial expression labels $l \in \mathbf{R}^8$, we use a one-hot vector to label the eight emotion classes. We also implemented all four networks: $N_1$, $N_2$, $N_3$ and the CAAE-like structure. Parameter settings were the same as the previous experiment. The images generated by the networks are shown in Fig. 2(b). In addition to the previous conclusion that $N_2$ better preserves identity and generates more realistic expressions than the others, we find that extreme expression changes tend to cause more artifacts when the feedforward connection is between the high-resolution layers. In the last row of Fig. 2(b), the generation of faces by $N_1$ only works well for subtle facial changes. Dramatic expression transitions, like from happiness to surprise, causes ghost artifacts and nonrealistic image synthesis. Fig. 5 shows further synthesis results from N2 and demonstrates its ability to interpolate between emotions.

### 3.3. User Study

To quantify human perception of the ability of the CDAAE network to preserve identity and to generate quality expressions, we created an online survey using both the DISFA and RaFD datasets. The survey is separated into three sections. In the first section, there were 33 questions regarding identity recognition on the RaFD dataset. We randomly selected 33 source images of different subjects with arbitrary expressions, each with a random target expression label, to generate faces using the $N_2$, $N_3$ and CAAE models. The source image subjects were unseen during training. We presented four images to the users: an actual image from the dataset with the target expression label and images generated by the three structures in random order. Users were asked to select which of the three randomly ordered images best corresponded to the actual image. Subjects were also free to indicate none of them. In the second section, the same test was conducted on seven source images from the DISFA database.

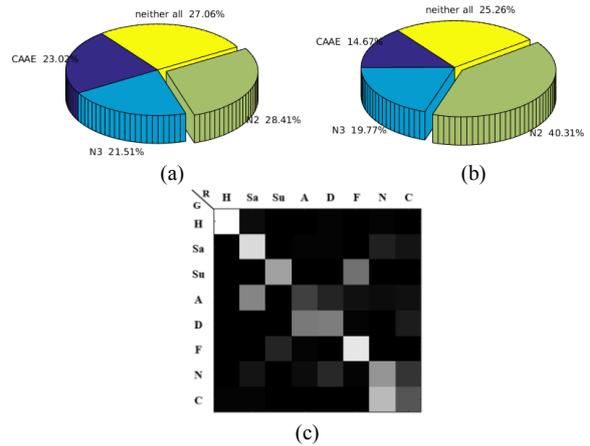

Fig.6. The statistical results of our user study. (a) RaFD user preference. Users preferred the images generated by the N2 structure. (b) DISFA user preference. N2 is preferred by a higher percentage than RaFD. These result shows that the CDAAE has advantages with smaller databases. (c) The confusion matrix of facial recognition from the users. Users can generally recognize most of the generated expressions, although it is still hard for them to differentiate neutral and contempt, anger and disgust, fear and surprise. (**G**:ground, **R**:recognition)

The third section included 16 questions for the RaFD database. We randomly picked two generated faces for each emotion class, and asked the users to identify the facial expression.

In the end, we received feedback from 112 volunteers randomly recruited online, and obtained 3696 votes for section one, 784 votes for section two, and 1792 votes for section three. The statistical results are shown in Fig.6. Our survey has several findings. First, the proposed generative models efficiently preserve identity information. 72.94% and 74.74% of users thought that at least one of the generated images was similar to the actual images from the RaFD and DISFA datasets respectively. Users preferred the images generated by the CDAAE ($N_2$) network more often other images generated by the other networks for both databases. This suggests that the CDAAE ($N_2$) has a better ability to preserve identity on DISFA, even though the number of subjects in the training datasets is small, and the facial changes are not too extreme. Second, the CDAAE successfully generates target facial expressions on unseen faces. On average 55.75% of the users recognized the ground truth expression of the target face. The recognition rate is highest for happiness (88.39%), sadness(75%) and fear(79.91%). Some expression pairs (neutral/contempt, anger/ disgust, fear/ surprise) are hard to differentiate.

### 4. Conclusion

In this paper, we have proposed the Conditional Difference Adversarial Autoencoder (CDAAE) for facial expression synthesis on new faces. Given one query face with a random facial expression and a target label (AU label or emotion label), this network generates a facial image of the same subject with an arbitrary target facial expression while greatly preserving identity information. This is achieved by adding a feedforward connection, which enables the reuse of encoded low-level features during the decoding process in a Conditional Adversarial Autoencoder. This frees up the higher levels of the network to concentrate on encoding differences between pairs of facial expressions. Note that the facial expression of the input image need not be neutral or even known. Only the output facial expression needs to be specified explicitly. We investigated the performance of different locations of this feedforward connection, and the effect of deleting it entirely. Our results show that the N2 structure does well both in preserving identity information and in synthesizing realistic facial expressions. We demonstrated this qualitatively by visualization and quantitatively by a user study. In summary, our model can generate photorealistic facial expressions. When used with the FACS, it can be used to generate arbitrary expressions through combinations of the AUs. In addition, it can be used to interpolate unseen expressions. These advances enable the CDAAE to achieve high quality results on database with a relatively small number of subjects. We are currently investigating how this method can be further applied to data augmentation for facial expression recognition and AU analysis.